\documentclass[conference]{IEEEtran}
\IEEEoverridecommandlockouts    % to override locked commands

\usepackage{multirow}
\usepackage{lipsum}
\usepackage{amsmath}
\usepackage{amssymb}
\usepackage[ruled,vlined]{algorithm2e}
\usepackage{graphicx}
\graphicspath{{./Figures/}}
\usepackage{xcolor}

\definecolor{darkgreen}{rgb}{0,0.5,0}
\definecolor{darkred}{rgb}{0.6,0,0}
\definecolor{purple}{rgb}{0.5,0,0.5}
\definecolor{orange}{rgb}{1,0.4,0}

\newcount\Comments
\newcommand{\ifshowcomments}[2]{%
  \ifnum\Comments=0
    #1% show
  \else
    #2% hide
  \fi
}

\newcommand{\robab}[1]{\ifshowcomments{\textcolor{blue}{[Robab: #1]}}{}}
\newcommand{\xingyu}[1]{\ifshowcomments{\textcolor{darkgreen}{[Xingyu: #1]}}{}}

\newcommand{\jerein}[1]{\ifshowcomments{\textcolor{orange}{[Jerein: #1]}}{}}

\usepackage{tabularx}   % better column wrapping
\usepackage{booktabs}
\newcolumntype{L}[1]{>{\raggedright\arraybackslash}p{#1}}

\usepackage{makecell}

\newtheorem{theorem}{Theorem}

\newtheorem{definition}{Definition}

\usepackage[hidelinks]{hyperref}   % or [colorlinks=true,linkcolor=black,citecolor=blue,urlcolor=blue]
\usepackage{array}       % for >{\raggedright\arraybackslash}

\usepackage{pifont}
\usepackage{comment}

\usepackage[normalem]{ulem}

\newcommand{\add}[1]{\textcolor{black}{#1}}

\title{\LARGE \bf Uncertainty-Aware Measurement of Scenario Suite Representativeness for Autonomous Systems %\xingyu{no need to mention the method, as this is a very applied conference and they want to focus on the problem, rather than the method, IMHO. In the intro, after we frame the problem properly then we can explain why bayesian + imprecise prob. is the right method to apply}\robab{Sure!}
}

\author{
Robab Aghazadeh-Chakherlou, Siddartha Khastgir, Xingyu Zhao, Jerein Jeyachandran, and Shufeng Chen\\
WMG, University of Warwick, Coventry, UK\\
\{r.aghazadeh-chakherlou, s.khastgir.1, xingyu.zhao, jerein.jeyachandran, shufeng.chen.1\}@warwick.ac.uk
}

\begin{document}
	
	\maketitle
	\thispagestyle{empty}
	\pagestyle{empty}
	
	%%%%%%%%%%%%%%%%%%%%%%%%%%%%%%%%%%%%%%%%%%%%%%%%%%%%%%%%%%%%%%%%%%
    %\begin{comment}
	\begin{abstract}   
    Assuring the trustworthiness and safety of AI systems, e.g., autonomous vehicles (AV), depends critically on the data-related safety properties, e.g., representativeness and completeness of the datasets used for their training and testing. Among these properties, this paper focuses on representativeness---the extent to which the scenario-based data used for training and testing, reflect the operational conditions that the system is designed to operate safely in, i.e., Operational Design Domain (ODD) or expected to encounter, i.e., Target Operational Domain (TOD). \jerein{To align with OASISS definition, can we rephrase to: "reflect the real-world operational conditions that the system is expected to encounter i.e., Target Operational Domain (TOD)." As representativeness is the extent to which the dataset reflects the TOD. E.g., if rainfall is not part of the ODD, but it is part of the TOD, we check for representativeness. So it is independent of the ODD of the system. This distinction is important only in the completeness evaluation of OASISS.}\robab{I have mentioned some studies in introduction that they are using the ODD as comparison reference, and I think we need t let the reader to know that either ODD or TOD can be as comparison reference. So, I decided to keep it.}We propose a probabilistic method that quantifies representativeness by comparing the statistical distribution of features encoded by the scenario suites with the corresponding distribution of features representing the TOD, acknowledging that the true TOD distribution is \textit{unknown}, as it can only be inferred from limited data. 
We apply an imprecise Bayesian method to handle limited data and uncertain priors. The imprecise Bayesian formulation produces interval-valued, uncertainty-aware estimates of representativeness, rather than a single value. 
We present numerical examples comparing the distributions of the scenario suite and the inferred TOD across operational categories (e.g., weather and road type) under dependencies and prior uncertainty. We estimate representativeness locally (between categories) and globally as an interval. 
\end{abstract}
%\end{comment}	
	%%%%%%%%%%%%%%%%%%%%%%%%%%%%%%%%%%%%%%%%%%%%%%%%%%%%%%%%%%%%%%%%%%
	\section{Introduction}
	\label{sec_introduction}
The rapid integration of Artificial Intelligence (AI) into autonomous systems has enabled machines to perform increasingly complex and safety-critical functions that were traditionally carried out by humans. In the context of autonomous vehicles and other intelligent systems, this transformation raises a pressing challenge: ensuring that the data-related safety properties governing AI-based decision-making components are sufficient to enable safe and reliable operation. \xingyu{guarantee seems to be a very strong word, I would say "enable"} \robab{you are right. Done!}

%Table \ref{tab_data-safety-props} summarises the data-related safety properties defined in PD ISO/PAS 8800:2024 \cite{ISO8800_2024}, DATA SAFETY (V3.6) \cite{SCSC_DataSafety_2024}, and ISO/IEC 5259-1 \cite{ISO5259_2025}, which together specify the key dataset qualities required for AI and autonomous-system safety assurance.
Table~\ref{tab_data-safety-props} summarises data-related safety properties defined in PD ISO/PAS~8800:2024~\cite{ISO8800_2024}, DATA~SAFETY~(V3.6)~\cite{SCSC_DataSafety_2024}, and ISO/IEC~5259-1~\cite{ISO5259_2025}, which together specify key dataset qualities for AI and autonomous-system safety assurance.

\begin{table}[h!]
\centering
\caption{Data-related safety properties from PD ISO/PAS 8800:2024~\cite{ISO8800_2024}, DATA SAFETY (V3.6)~\cite{SCSC_DataSafety_2024}, and ISO/IEC 5259-1~\cite{ISO5259_2025}.}
\vspace{-6pt} % tighten space between caption and table
\label{tab_data-safety-props}
\footnotesize
\begin{tabular}{@{}p{2.1cm} p{6.2cm}@{}}
\toprule
\textbf{Source} & \textbf{Data-Related Safety Properties} \\
\midrule
\makecell[c]{\textbf{PD ISO/}\\[-2pt]\textbf{PAS 8800}~\cite{ISO8800_2024}} &
Accuracy; Completeness; Correctness / Fidelity; Independence of datasets; Integrity; Representativeness; Temporality; Traceability; Verifiability. \\
\midrule
\makecell[c]{\textbf{DATA}\\[-2pt]\textbf{SAFETY}~\cite{SCSC_DataSafety_2024}} &
Integrity; Completeness; Accuracy; Timeliness / Temporality; Verifiability; Fidelity / Representation; Lifetime; Confidentiality; Availability. \\
\midrule
\makecell[c]{\textbf{ISO/IEC}\\[-2pt]\textbf{5259-1}~\cite{ISO5259_2025}} &
Portability; Understandability; Auditability. \\
\bottomrule
\end{tabular}
\end{table}

While properties such as accuracy, completeness, integrity, and traceability (Table~\ref{tab_data-safety-props}) concern local or procedural aspects of data quality, representativeness uniquely addresses the global, distributional relationship between a dataset and the real-world conditions it is intended to model. Thus, among those properties, \textit{representativeness}—defined as \textit{``the distribution of data corresponds to the information in the environment of the phenomenon to be captured; it is free of biases''}~\cite{ISO8800_2024}—is of particular importance. %It determines whether the data used to develop and evaluate an AI model are truly sufficient to reflect the conditions the system will encounter in the real world.

%In existing literature and standards, the terms environment and dataset carry related but distinct meanings. Their interpretation varies across studies—from operational design domain (ODD) distributions~\cite{langner_statistical_2023}, to observed real-world test data (test sets)~\cite{de_scenario_2022}, and multi-source reference statistics such as maps, weather, and traffic databases~\cite{hiller_creating_2022}. 
In existing literature and standards, the term environment carries related but distinct meanings. The notion of environment ranges from Operational Design Domain (ODD) distributions~\cite{langner_statistical_2023} to Target Operational Domain (TOD) specifications~\cite{ISO34503_2023}.
\add{ISO~34503~\cite{ISO34503_2023} distinguishes between ODD, which defines the conditions under which an ADS is designed to operate safely, and TOD, which describes the real-world conditions the ADS may encounter and must handle safely. The standard allows different relationships between these domains, including partial overlap or one being broader than the other\footnote{In practice, the TOD is often broader than the ODD.}. 
We %focus on the common case where the TOD is broader than the ODD, and 
restrict representativeness assessment to the intersection $\mathrm{TOD} \cap \mathrm{ODD}$, considering only ODD-compliant scenarios. Representativeness is therefore evaluated by comparing the statistical distribution of features (i.e., scenario attributes such as road, weather, and traffic conditions) in the scenario suites with the corresponding distribution of features representing this restricted TOD\footnote{For brevity, we will refer to them as the ``scenario suite distribution'' and the ``TOD distribution''.}.
}

%According to ISO 34503 \cite{ISO34503_2023}, the TOD defines the real-world conditions that an autonomous driving system (ADS) may experience and is required to safely operate in, whereas the Operational Design Domain (ODD) specifies the conditions under which the ADS is designed to operate safely. In practice, the TOD generally represents a broader and more variable environment than the ODD. Therefore, any discrepancy between the ODD and TOD—often referred to as the ODD–TOD gap—creates uncertainty regarding whether the available data sufficiently represent the operational conditions the system will encounter once deployed. In practice, this necessitates assessing the adequacy of the training and testing datasets—whether they are complete and representative of the operational conditions for which the system is intended. However, current approaches lack quantitative, model-agnostic mechanisms to perform this evaluation.

%Despite the growing availability of standards such as ISO~34503~\cite{ISO34503_2023} and ISO/PAS~8800~\cite{ISO8800_2024}, as well as safety assurance frameworks such as AMLAS~\cite{hawkins_guidance_2021}, which primarily emphasize \emph{process-level assurance} and \emph{qualitative reasoning}, only a few studies such as~\cite{jeyachandran_OASISS_2025, langner_statistical_2023} have attempted to \emph{quantify} data-related safety properties—such as representativeness or completeness—and these remain exceptions rather than the norm.

In the context of AI safety assurance, and following ISO/PAS 8800 \cite{ISO8800_2024}, the term data encompasses all information used to train, test, and validate AI components. In the case of autonomous systems, this data is typically scenario-based, describing structured representations of the situations the system may encounter during operation. A study \cite{jeyachandran_OASISS_2025} interprets this concept operationally by treating structured scenarios as the fundamental data that may be used to train and test AI models (scenario suites). \jerein{in a broad sense, individual scenarios need not always be designed represent the TOD. So can we rephrase as: "data that may be used to train and test AI models (scenario suites)."}\robab{Done!}Each scenario captures a combination of operational variables such as weather, road type, illumination, etc.
%, thereby sampling the multidimensional space of the TOD.
Assessing AI system safety involves evaluating data-related safety properties (Table \ref{tab_data-safety-props}), which together determine the adequacy of the dataset. Among the key adequacy properties, this paper focuses on representativeness.

%Following the definition of representativeness in ISO~8800~\cite{ISO8800_2024} and the tailored definition of data introduced in OASISS~\cite{jeyachandran_OASISS_2025}, we define representativeness as the degree to which the distribution of data used for testing or validation reflects the conditions expected within the TOD.
%In this context, data refer to the set of \emph{test scenarios}—each characterized by specific operational factors such as weather, road type, traffic, time of day etc.,—that collectively describe the operational conditions under which the system is evaluated.

%From a methodological standpoint, representativeness provides a measurable bridge between assurance theory and probabilistic modelling: it can be formally expressed as a \emph{distributional discrepancy} between the empirical scenario suite and the true (but unknown\xingyu{unknown? or never known with certainty}) TOD distribution.
\xingyu{Not sure at this stage, if the reader can understand why TOD distribution is unknown or uncertain. Maybe it is useful to say something like:... true (but uncertain) TOD distribution (It is unknown because blah blah )}\robab{Done! Please see the following paragraph.}

As mentioned earlier, representativeness %provides a measurable bridge between assurance theory and probabilistic modelling: it 
can be formally expressed as a \emph{distributional discrepancy} between the empirical scenario suite and the true but unknown TOD distribution.
The TOD distribution is unknown because it represents the real-world operational conditions that the system will encounter, which can only be estimated from limited, potentially biased, or evolving observational data.

%OASISS~\cite{jeyachandran_OASISS_2025} introduces an approach to quantitatively assess the adequacy of AI training and testing datasets relative to the targeted operational conditions. Building on the definitions of completeness and representativeness from ISO/PAS 8800 \cite{ISO8800_2024}, OASISS focuses on the completeness property—defined as the coverage of expected and out-of-ODD conditions—and provides numerical scores to quantify this aspect. Representativeness, in contrast, is adopted conceptually from ISO 8800 as the degree of alignment between the dataset distribution and the real-world conditions in the Target Operational Domain (TOD), but is not explicitly quantified in OASISS.  

\begin{figure*}[t]
    \centering
    \includegraphics[width=0.75\textwidth]{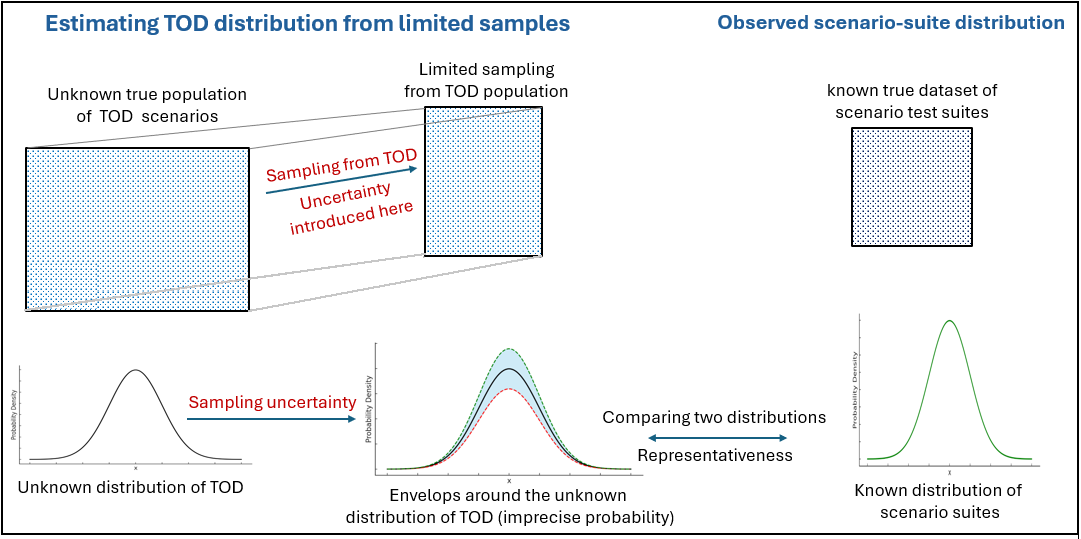}
    \caption{Schematic illustration of the representativeness evaluation process. The true population of TOD scenarios is unknown, and sampling from it introduces uncertainty, represented by imprecise probability envelopes around the estimated distribution. Representativeness is then assessed by comparing this uncertain TOD distribution with the known distribution of the  scenario test suite. \xingyu{in the top-left fig, the sampling process is a not misleading... now it looks like you are reusing a specific part of the population data rather than random sampling from the whole population... maybe a quick call can clarify it better.. also, I think labeling desctiptive statistics and statstical inferenec in the fig would help}\robab{Done!}}
    \vspace{-6pt} % tighten space between caption and figure
    \label{fig_represent_process}
\end{figure*}

A key challenge lies in the multidimensionality of the TOD, which includes variables such as weather, road type, traffic density, and lighting, each spanning multiple levels (e.g., rainy/clear, urban/rural). %In this study, we define each \emph{category} as a unique combination of these variable levels.
Estimating a joint distribution from limited observations is difficult. %as it reflects a population-level process with inherent uncertainty.
If the occurrence frequencies of each category (e.g., clear–rural–light traffic–day, rainy–rural–light traffic–day, etc.) were known with certainty, one could deterministically fit a distribution to the data. In practice, however, such knowledge is rarely available, motivating a probabilistic approach that assigns distributions over category occurrences rather than fixed values (l.h.s of Fig.~\ref{fig_represent_process}).

In contrast, estimating the empirical distribution of the scenario suites used is relatively straightforward, as these data are directly available. This corresponds to a descriptive statistical analysis of the observed dataset, rather than statistical inference about an unknown population (r.h.s of Fig.~\ref{fig_represent_process}).
\xingyu{This is the typical example of descriptive statstics of a given dataset vs statistical inference of a unknown population; not sure if we should mention this two types of methods at all... maybe a bit too much for IV readers?}\robab{Done!}

The central contribution of this work is therefore the estimation of the TOD joint distribution using a Bayesian framework (statistical inference, shown on the l.h.s of Fig.~\ref{fig_represent_process}) well-suited for reasoning under data sparsity and uncertainty. Furthermore, we employ imprecise probability to represent uncertainty in the prior distribution used in the Bayesian inference. In general, our work formalizes representativeness as a \emph{distributional discrepancy} between the empirical scenario suite and the \textit{``unknown''} TOD distribution, and by deriving a Bayesian framework to quantify it together with its epistemic uncertainty (Fig.~\ref{fig_represent_process}). Consequently, the resulting discrepancy metric is expressed as an uncertainty-aware estimate interval. %Figure~\ref{fig_represent_process} illustrates the representativeness evaluation process used in this paper.

There are different statistical metrics available for comparing probability distributions. In this work, we employ Total Variation Distance (TVD) and Jensen–Shannon Divergence (JSD) for this purpose.

The remainder of this paper is organized as follows. Section~\ref{sec_related_work} reviews related work on dataset representativeness in AI safety assurance. Section~\ref{sec_metric_represent} presents the proposed metric for quantifying representativeness. A numerical example illustrating its application is provided in Sec.~\ref{sec_numerical_example}. Section~\ref{sec_discussion_conclusion} discusses the results, and finally, Sec.~\ref{sec_conclusion} concludes the paper.

	%%%%%%%%%%%%%%%%%%%%%%%%%%%%%%%%%%%%%%%%%%%%%%%%%%%%%%%%%%%%%%%%%%
\section{Related Work}
\label{sec_related_work}
This section reviews existing approaches to representativeness evaluation and imprecise probability to position the proposed framework.

\subsection{Comparison with Existing Methods on Representativeness}

Recent work in automated driving has recognized that the \emph{representativeness} of test scenarios is a key determinant of the credibility of safety evidence. 
Langner et al.~\cite{langner_statistical_2023} formalize representativeness as a \emph{statistical distance} between two empirical distributions: the observed \textit{open-road test data} and the \textit{target operational design domain (ODD)} reference statistics. 
%The comparison is therefore between measured test data and observed ODD statistics. 
De~Gelder et al.~\cite{de_scenario_2022} apply this idea to the comparison between generated scenario data and real-world driving data. 
They use a combination of singular value decomposition (SVD) and kernel density estimation (KDE) to model the joint distribution of driving parameters, and %employ the \emph{Wasserstein distance} to 
evaluate how closely the generated scenarios approximate the empirical distribution of real-world data. 
Hiller et al.~\cite{hiller_creating_2022} shift focus from the metric itself to the process of representativeness assessment. 
Their framework compares distributions of recorded driving data against reference statistics derived from heterogeneous sources such as OpenStreetMap, meteorological services, traffic counters, and accident databases. 
%This enables data-driven coverage evaluation, but assumes that all reference sources are accurate and complete. 
In \cite{jeyachandran_OASISS_2025}, representativeness is defined as the similarity between the empirical distribution of operational conditions in the scenario suite and the corresponding distribution in the TOD. Both are modeled as categorical probability distributions derived from ISO-standardized operational factors. %The Jensen–Shannon Divergence (and optionally the Total Variation Distance) is used as the quantitative metric, yielding a bounded measure of distributional alignment.
At the assurance level, the JRC report on Trustworthy Autonomous Vehicles~\cite{fernandez_trustworthy_2021} identifies representativeness and data coverage as measurable properties required for trustworthy validation and testing. 
While this establishes representativeness as a normative requirement, it does not specify how to quantify it or account for uncertainty.

The above methods evaluate representativeness as a distance between distributions of two \emph{empirical} datasets: generated versus real-world \cite{de_scenario_2022}, test versus reference ODD data \cite{langner_statistical_2023}, or scenario suites versus TOD \cite{jeyachandran_OASISS_2025}. The real-world dataset is treated as a reliable surrogate for the true operational domain, assuming that sufficient data exist to capture its variability. 
In practice, however, this assumption may not hold when real world data are sparse or partially observed. Some studies, such as Gelder et al~\cite{de_scenario_2022}, highlight this issue and quantify how sampling bias in real-world testing affects the validity of validation results using frequency-based divergence measures.  %our approach recognizes that such references may be uncertain, conflicting, or geographically limited, and therefore must be represented probabilistically rather than deterministically.
Our probabilistic framework (Bayesian inference) addresses this by comparing the empirical distribution of the scenario suite with an estimated approximation of the true, but \textit{unknown}, TOD distribution. Instead of assuming that the TOD can be directly estimated from abundant real-world data, we treat its distribution as uncertain and propagate epistemic uncertainty into the representativeness metrics. 
 Thus, our work extends deterministic distribution-matching approaches into a robust, uncertainty-aware framework.

\subsection{Imprecise Probability}

While classical Bayesian methods provide a coherent mechanism for uncertainty propagation, they depend on the specification of a single prior distribution, which can be challenging to justify under sparse data. 
\emph{Imprecise probability} frameworks~\cite{augustin2014introduction}
%\cite{augustin2014introduction,imprecision_luck_walter_2009,zhao_interval_2020,zhao2024bayesian}
%\cite{augustin2014introduction,troffaes2007decision,utkin_imprecise_2018,imprecision_luck_walter_2009,zhao_interval_2020,zhao2024bayesian}
relax this assumption by representing uncertainty using \emph{set} of distributions (credal sets) instead of a single one (Table~\ref{tab_ip_vs_classical}). 
%Such approaches generalize Bayesian inference by producing interval-valued or set-valued posterior quantities, offering robustness against prior misspecification. 
%When using imprecise probability, 
Estimating such \emph{set} of posterior distributions %rather than a single posterior, which 
poses significant computational challenges. This is because there is generally no monotonic relationship between the prior and the posterior. However, when the prior belongs to the exponential family, a reparameterization can be applied---expressing the prior in terms of its \emph{strength} and \emph{mean} parameters---according to the iLUCK model proposed by~\cite{imprecision_luck_walter_2009} and applied in different studies (\cite{zhao_interval_2020,zhao2024bayesian}). In the context of our categorical
data, the iLUCK model~\cite{imprecision_luck_walter_2009} allows the specification of bounds on these parameters. This formulation preserves conjugacy while enabling monotonic propagation of prior imprecision into posterior intervals. This monotonicity property allows us to compute the posterior bounds by evaluating only the corner cases of the prior parameter space, rather than calculating the posterior for every possible prior configuration. 

\begin{table}[t]
\centering
\caption{Comparison of Classical Bayesian and Imprecise Probability for a coin-flip example.}
\label{tab_ip_vs_classical}
\setlength{\tabcolsep}{4pt}
\renewcommand{\arraystretch}{1.1}
\footnotesize
\begin{tabular}{@{}lcc@{}}
\toprule
\textbf{Aspect} & \textbf{Classical Bayesian} & \textbf{Imprecise Probability} \\
\midrule
Prior &
$\mathrm{Beta}(3,3)$ &
$\mathrm{Beta}(\alpha,\beta),$ $\alpha,\beta\!\in\![2,4]$ \\
Posterior &
$\mathrm{Beta}(8,10)$ &
$\mathrm{Beta}(5{+}\alpha,7{+}\beta)$, $\alpha,\beta\!\in\![2,4]$ \\
Posterior mean &
$\mathbb{E}[\theta \mid D] = 0.44$ &
$[0.39,\,0.48]$ \\
\bottomrule
\end{tabular}
\end{table}

\xingyu{Similar to our HIP-LLM paper, I wonder if a quick example of iLUCK can help the reader understand.. If yes, maybe we can just reuse what is in the HIP-LLM paper} \robab{Done!}

\section{A representativeness metric}%XZ: like Lorenzo used to tell me, metrics by default are quantitatve, so no need to say quantitative metric...
\label{sec_metric_represent}

\begin{comment}
Borrowing the definition of data representativeness from relevant standards, in this work it is defined as the degree to which a dataset or scenario suite accurately reflects the Target Operational Domain (TOD).
 Let us to define the TOD precisely in our problem:
 \begin{definition}
     The \emph{Target Operational Domain} (TOD) defines the real-world conditions under which the system is intended to operate safely. It characterizes the relevant operational factors/variables that jointly describe the system’s deployment environment and use conditions. These factors can be grouped into three complementary dimensions: the \emph{environment}, indicating \emph{where} the system operates (e.g., road type, weather, illumination, traffic conditions); the \emph{task}, describing \emph{what} the system is performing (e.g., lane keeping, overtaking, pedestrian detection); and the \emph{context}, referring to \emph{under what circumstances or constraints} the operation takes place (e.g., time of day, regulatory conditions, or interaction with other road users). Together, these dimensions define the space of real-world situations that the system may encounter once deployed.
 \end{definition}
\end{comment} 
Given the definition of TOD and tailored definition of data in this study, it is conceived as a multidimensional space that describes the range of real-world operating conditions under which an AI system is expected to function safely.
Each dimension of this space corresponds to an operational variable, such as weather, road type, illumination, etc., (Table~\ref{tab_category}).

Because many of these operational variables are continuous in nature---such as illumination intensity, temperature, or speed---it is often \emph{easier and more practical} to represent the TOD in a finite, analyzable form through \emph{discretization} (see Table~\ref{tab_category}).
In practice, standards and safety guidelines such as ISO/PAS~8800:2024~\cite{ISO8800_2024} and DATA~SAFETY~(V3.6)~\cite{SCSC_DataSafety_2024} establish the requirement for a well-defined coverage of the operational domain but leave the specific discretization boundaries to be determined by domain-specific standards or, in their absence, through expert judgment. %\robab{check using hat for the parameters in the document.}

\begin{table}[h!]
\centering
\caption{Illustrative hierarchy of variables and categories in TOD.}
\vspace{-6pt} % tighten space between caption and table
\label{tab_category}
\footnotesize
\begin{tabular}{@{}p{1cm} p{3cm} p{3.9cm}@{}}
\toprule
\makecell{\textbf{Hierarchy}\\ \textbf{level}} & \textbf{Example elements} & \textbf{Interpretation / role} \\ 
\midrule
\makecell{\textbf{Variables/} \\ \textbf{Factors}} &
\makecell{Weather, Road type, Time\\ of day, Traffic density,\\ Speed band, etc.} &
\makecell{Dimensions describing operational\\ and environmental conditions.\\ Each variable may be continuous\\ or discrete.} \\
\midrule
\makecell{\textbf{Discretized} \\ \textbf{levels}} &
\begin{tabular}[c]{@{}l@{}}Weather: \{Clear, Rain, Fog\}\\
\makecell[l]{Road type:\\ \{Highway, Urban, Rural\}}\\
Time: \{Day, Dusk, Night\}\end{tabular} &
\makecell{Continuous variables are\\ partitioned into discrete bins\\ (e.g., speed ranges) to enable\\ categorical representation.} \\
\midrule
\makecell{\textbf{Joint}\\ \textbf{categories}} &
\begin{tabular}[c]{@{}l@{}}(Rainy, Urban, Night)\\
(Clear, Highway, Day)\\
(Foggy, Rural, Dusk), etc.\end{tabular} &
\makecell{Each unique combination of\\ discretized variable levels forms\\ one category \(k \in \{1,\ldots,K\}\),\\ representing a distinct operational\\ condition in the TOD.} \\
\bottomrule
\end{tabular}
\end{table}

Given these joint categories of variables, our goal is to compare the estimated distributions of the TOD and the scenario suite to evaluate representativeness. In the first step, we sample from the TOD population and estimate its distribution using an imprecise Bayesian framework, as formulated in Theorem~\ref{thm_TOD_posterior}. Then in Sec.~\ref{sec_metric} a metric is presented to quantify the representativeness.

\subsection{Estimating the Posterior Distribution Envelopes of TOD}
\label{sec_post_TOD}
Theorem~\ref{thm_TOD_posterior} formalizes the computation of lower and upper bounds on the TOD posterior under imprecise priors.
 
\begin{theorem}[Posterior inference for the TOD distribution]
\label{thm_TOD_posterior}
Let the Target Operational Domain (TOD) be represented by a categorical random variable 
$Z \in \{1,\ldots,K\}$ with unknown probability vector 
$\boldsymbol{\theta}_{\text{TOD}} = (\theta_1,\ldots,\theta_K) \in \Delta_{K-1}$ where, $\Delta_{K-1}$ is all valid probability distributions over $K$ categories:
\[\Delta_{K-1} = \left\{ 
\boldsymbol{\theta} = (\theta_1, \ldots, \theta_K) \in \mathbb{R}^K :
\theta_k \ge 0,\; \sum_{k=1}^K \theta_k = 1
\right\}\]

Given observed category counts $\mathbf{k}=(k_1,\ldots,k_K)$ from $n$ i.i.d.\footnote{\add{i.i.d. means that each scenario is treated as an independent draw from the TOD distribution, so the outcome of one scenario does not affect another.}}\ samples,
the likelihood is multinomial:
\[
f(\mathbf{k}\mid \boldsymbol{\theta}_{\text{TOD}})\propto \prod_{k=1}^K \theta_k^{k_k}.
\]

Assuming a Dirichlet prior with strength $n^{(0)}$ and prior mean vector $\boldsymbol{y}^{(0)}$, the posterior is
\[
\boldsymbol{\theta}_{\text{TOD}}\mid \mathbf{k} 
\sim 
\mathrm{Dirichlet}\!\big(n^{(0)}\boldsymbol{y}^{(0)}+\mathbf{k}\big).
\]

To account for prior uncertainty, let 
$\boldsymbol{y}^{(0)} \in \mathcal{Y}_{\text{TOD}}\subseteq\Delta_{K-1}$ (where, $\mathcal{Y}_{\text{TOD}}$ is the set of all possible prior mean vectors)
and $n^{(0)}\in[\underline{n},\overline{n}]$.
Then the \emph{posterior credal set} is
\begin{align}
    &\mathcal{P}(\boldsymbol{\theta}_{\text{TOD}}\mid\mathbf{k})
= \nonumber \\
& \hspace{0.5cm}\Big\{
\mathrm{Dirichlet}\!\big(n^{(0)}\boldsymbol{y}^{(0)}+\mathbf{k}\big)
:\;
\boldsymbol{y}^{(0)}\in\mathcal{Y}_{\text{TOD}},
\;
n^{(0)}\in[\underline{n},\overline{n}]
\Big\} \nonumber
\end{align}

The joint posterior forms a family of Dirichlet distributions over the simplex, whose lower and upper envelopes are attained at the extreme points of $\mathcal{Y}_{\text{TOD}}$ and $[\underline{n}, \overline{n}]$. Marginally, each $\theta_k$ follows a corresponding family of Beta\footnote{\add{This is the mathematical property of the Dirichlet.}} distributions, with posterior bounds obtained in the same way.

%The marginal posterior of each component $\theta_k$ forms a family of Beta distributions, and the\emph{lower} and \emph{upper} posterior cumulative distributions are obtained as the pointwise infimum and supremum over this family, corresponding to the extreme points of $\mathcal{Y}_{\text{TOD}}$ and $\{\,\underline{n},\overline{n}\,\}$.
\end{theorem}

See Appendix~\ref{proof_outline_appendix} for the outline of proof.

\subsection{Representativeness Metric under Uncertainty}
\label{sec_metric}
%Having established a method to infer the posterior distribution of the TOD in Theorem~\ref{thm_TOD_posterior}, we now address how to quantify the representativeness of the scenario suite by measuring the discrepancy between two distributions: the empirical scenario suite distribution $\hat{\boldsymbol{\pi}}_{\mathrm{suite}}$ and the posterior TOD distribution $\boldsymbol{\theta}_{\mathrm{TOD}}|{\bf k}$.

Given the posterior TOD distribution from Theorem~\ref{thm_TOD_posterior}, representativeness is quantified as the discrepancy between $\hat{\boldsymbol{\pi}}$ (empirical scenario suite distribution) and $\boldsymbol{\theta}_{\mathrm{TOD}}\mid\mathbf{k}$. Representativeness is quantified by the alignment between these two categorical distributions over the same $K$ joint categories, using two complementary measures from information and probability theory:

%A representativeness metric must capture the overall alignment between these two categorical distributions defined over the same $K$ joint categories. We employ two complementary measures that are well-established in information theory and probability theory:

\subsubsection{Total Variation Distance (TVD)}
TVD provides an intuitive measure of the maximum difference in probability that the two distributions assign to any event:
\begin{align}
\label{eq_TVD}
    D_{\mathrm{TV}}(\hat{\boldsymbol{\pi}}_{\mathrm{suite}}, \boldsymbol{\theta}_{\mathrm{TOD}}) = \frac{1}{2}\sum_{k=1}^{K}|\hat{\pi}_k - \theta_k| 
\end{align}

The factor $\frac{1}{2}$ ensures that $D_{\mathrm{TV}} \in [0, 1]$, where 0 indicates perfect alignment and 1 indicates complete disjointness. The TVD has a clear operational interpretation: it represents the maximum difference in probability that the two distributions assign to any subset of categories. A value of $D_{\mathrm{TV}} = 0.05$, for example, means that the two distributions differ by at most 5\% in their probability assignments to any collection of operational conditions.

\subsubsection{Jensen-Shannon Divergence (JSD)}
JSD is a symmetrized and smoothed version of the Kullback-Leibler divergence, defined as:
\begin{align}
\label{eq_JS}
    & D_{\mathrm{JS}}(\hat{\boldsymbol{\pi}}_{\mathrm{suite}}, \boldsymbol{\theta}_{\mathrm{TOD}}) = \nonumber \\
    & \hspace{0.5 cm}\frac{1}{2}D_{\mathrm{KL}}(\hat{\boldsymbol{\pi}}_{\mathrm{suite}} \| {\bf m}) + \frac{1}{2}D_{\mathrm{KL}}(\boldsymbol{\theta}_{\mathrm{TOD}} \| {\bf m}) ,
\end{align}
where ${\bf m} = \frac{1}{2}(\hat{\boldsymbol{\pi}}_{\mathrm{suite}} + \boldsymbol{\theta}_{\mathrm{TOD}})$ is the average distribution, and $D_{\mathrm{KL}}$ denotes the Kullback-Leibler divergence:
\[D_{\mathrm{KL}}({\bf p} \| {\bf q}) = \sum_{k=1}^{K}p_k \log\frac{p_k}{q_k}, \qquad p = \hat{\pi}_{\mathrm{suite}}, \quad q = \boldsymbol{\theta}_{\mathrm{TOD}}\]

%Unlike the KL divergence, the JSD is symmetric, always finite (even when distributions have non-overlapping support), and bounded: $D_{\mathrm{JS}} \in [0, \log 2]$. The square root of JSD, $\sqrt{D_{\mathrm{JS}}}$, forms a proper metric and is particularly sensitive to differences in the bulk of the distributions rather than just the tails.

\subsubsection{Propagating Epistemic Uncertainty}

When using an imprecise prior, the posterior TOD distribution is represented by a \emph{credal set} rather than a single distribution. Consequently, both $D_{\mathrm{TV}}$ and $D_{\mathrm{JS}}$ become interval-valued:
\begin{align}
D_{\mathrm{TV}} &\in \left[\underline{D}_{\mathrm{TV}}, \overline{D}_{\mathrm{TV}}\right]\nonumber \qquad
D_{\mathrm{JS}} \in \left[\underline{D}_{\mathrm{JS}}, \overline{D}_{\mathrm{JS}}\right] \nonumber
\end{align}
where the lower and upper bounds are obtained by evaluating the discrepancy measure at the boundary values of the prior strength parameter $n^{(0)} \in [\underline{n}, \overline{n}]$. 
Because the posterior mean $\boldsymbol{\theta}_{\mathrm{TOD}}$ varies monotonically with the prior strength $n^{(0)}$, 
and both $D_{\mathrm{TV}}$ and $D_{\mathrm{JS}}$ are continuous in $\boldsymbol{\theta}_{\mathrm{TOD}}$, 
the minimum and maximum discrepancy values over the prior interval $[\underline{n}, \overline{n}]$ occur at the boundary points of this interval.

These bounds illustrate epistemic uncertainty in representativeness, indicating how sensitive the assessment is to prior beliefs about the TOD.
%Narrow intervals imply robustness, while 
Wide intervals suggest the need for more operational data to refine the estimate.

%For practical safety assurance, the interpretative thresholds of the Total Variation Distance (\(D_{\mathrm{TV}}\)) can be used to assess dataset representativeness relative to the Target Operational Domain (TOD). 
%These thresholds should be interpreted in the context of the application domain and the potential consequences of misrepresentation. 
%In safety-critical systems, even values within the “good” range may warrant further examination if they correspond to under-representation of high-risk scenarios.

\section{Numerical Example}
\label{sec_numerical_example}
%This section presents numerical examples illustrating the proposed method. 
\add{The numerical example presented in this section is intentionally simplified and uses a small number of abstract categorical variables to illustrate the proposed method in a transparent and reproducible manner.}%, rather than to model a complete real-world operational domain.}

\subsection{Problem Setting: Common Cases}

The TOD and the scenario suite share the same variable set, discretization, and categorical space. Both are defined over five operational factors: \(
W,R,D,T,S,
\)
where, Weather $W \in \{\text{Clear}, \text{Adverse}\}$, Road $R \in \{\text{Highway}, \text{Urban}\}$, Time $D \in \{\text{Day}, \text{Night}\}$, Traffic $T \in \{\text{Light}, \text{Heavy}\}$, and Speed $S \in \{\le 50,\; >50\}$.
\begin{comment}
\begin{itemize}
    \item Weather $W \in \{\text{Clear}, \text{Adverse}\}$,
    \item Road $R \in \{\text{Highway}, \text{Urban}\}$,
    \item Time $D \in \{\text{Day}, \text{Night}\}$,
    \item Traffic $T \in \{\text{Light}, \text{Heavy}\}$,
    \item Speed $S \in \{\le 50,\; >50\}$.
\end{itemize}
\end{comment}
Each combination $(w,r,d,t,s)$ defines one of $K = 2^5 = 32$ categories (Table~\ref{tab_cat32}), which serve as a common comparison basis for both the TOD and the scenario suite.  
\begin{table}[h!]
\centering
\caption{Mapping of joint categories to 5-bit codes.}
\vspace{-6pt} % tighten space between caption and table
\label{tab_cat32}
\begin{tabular}{cccccc}
\hline
\textbf{Code} & \textbf{Weather} & \textbf{Road} & \textbf{ToD} & \textbf{Traffic} & \textbf{Speed} \\
\hline
0  & Clear   & Highway & Day   & Light & \(\le 50\) \\
1  & Clear   & Highway & Day   & Light & \(>50\) \\
2  & Clear   & Highway & Day   & Heavy & \(\le 50\) \\
3  & Clear   & Highway & Day   & Heavy & \(>50\) \\
\(\vdots\) & \(\vdots\) & \(\vdots\) & \(\vdots\) & \(\vdots\) & \(\vdots\) \\
31 & Adverse & Urban   & Night & Heavy & \(>50\) \\
\hline
\end{tabular}
\end{table}

To ensure realism and consistency, dependencies among these factors are explicitly incorporated in both models through a shared Bayesian-network structure 
\((R \!\to\! W,\; D \!\to\! T,\; R \!\to\! S)\).  
This structure captures operationally relevant relationships such as weather conditions depending on road type, traffic patterns varying with time of day, and speed distributions differing across road environments. This structure is applied consistently to both the scenario suites and TOD.
It captures realistic relationships among scenario variables: road type influences both weather exposure and typical speed, while time of day affects traffic conditions.
The root marginal distributions are specified as
\begin{align}
    &P(R) = [0.60,\,0.40] \quad  (\text{Highway, Urban}), \nonumber \\
&P(D) = [0.65,\,0.35] \quad (\text{Day, Night}) \nonumber
\end{align}

The corresponding conditional probability tables (CPTs) are summarized in Table~\ref{tab_cpt_tod}.
\begin{table}[h!]
\centering
\caption{Conditional Probability Tables (dependent TOD prior).}
\vspace{-6pt}
\label{tab_cpt_tod}
\begin{tabular}{lcc}
\hline
\multicolumn{3}{l}{\textbf{$P(W \mid R)$ \; (Weather $\mid$ Road type)}}\\
\hline
 & Clear & Adverse \\
\hline
Highway & 0.75 & 0.25 \\
Urban   & 0.625 & 0.375 \\
\hline
\\[-0.9em]
\hline
\multicolumn{3}{l}{\textbf{$P(T \mid D)$ \; (Traffic $\mid$ Time of day)}}\\
\hline
 & Light & Heavy \\
\hline
Day   & 0.65 & 0.35 \\
Night & 0.50 & 0.50 \\
\hline
\\[-0.9em]
\hline
\multicolumn{3}{l}{\textbf{$P(S \mid R)$ \; (Speed $\mid$ Road type)}}\\
\hline
 & $\le 50$ & $> 50$ \\
\hline
Highway & 0.30 & 0.70 \\
Urban   & 0.65 & 0.35 \\
\hline
\end{tabular}
\end{table}

%Table~\ref{tab_cat32}  lists all combinations of the five factors and their corresponding codes.

\subsection{Problem Setting: Scenario Suite}
%The scenario suite was constructed to serve as an \emph{empirical test distribution} against which the inferred TOD distribution could be compared.
%Its purpose is to emulate the process of autonomous-vehicle scenario generation, where rare but safety-critical conditions are intentionally oversampled to ensure sufficient coverage for evaluation.
The scenario suite serves as an empirical test distribution for comparison with the inferred TOD distribution, emulating autonomous-vehicle scenario generation where rare but safety-critical conditions are intentionally oversampled to ensure sufficient evaluation coverage.

A synthetic dataset of \( n = 1{,}600 \) scenarios was generated by incorporating \emph{dependencies} among operational factors \((R \!\to\! W,\; D \!\to\! T,\; R \!\to\! S)\). 
%This structure captures realistic correlations observed in real-world environments: for instance, adverse weather may be more likely on rural highways, 
%nighttime driving tends to coincide with lower traffic density, and speed distributions differ across road types.
To reflect operational risk emphasis, higher sampling weights were assigned to more challenging conditions — %
\emph{adverse weather} (\(\times 1.4\)), \emph{urban roads} (\(\times 1.3\)), \emph{nighttime} (\(\times 1.25\)), 
\emph{heavy traffic} (\(\times 1.2\)), and \emph{low speeds} (\(\le 50~\mathrm{km/h}, \times 1.15\)).
Within each conditional relationship, multipliers were applied \emph{within parent contexts} and then renormalized to preserve valid conditional probability tables.
For example, in the conditional distribution \( P(W \mid R) \), the weight for adverse weather was increased by \(1.4\) relative to clear conditions, 
and the resulting probabilities were normalized within each road type.

The joint probability of each category \( c = (w, r, d, t, s) \) in the scenario suite was computed as
\begin{align}
&\hat{\pi}_{\text{suite}}(c)
 \;\propto\;
P(R{=}r) \, \times P(D{=}d) \times \nonumber \\
& \hspace{0.3 cm}\, P(W{=}w \mid R{=}r) \times
\, P(T{=}t \times \mid D{=}d) \times
\, P(S{=}s \mid R{=}r) \nonumber
\end{align}
%where each conditional term reflects both the dependency structure and the applied weighting multipliers.
%After normalization, the resulting probabilities formed a valid joint distribution:
where each conditional term reflects both the dependency structure and applied weighting multipliers. After normalization, the probabilities form a valid joint distribution:

\[
\hat{\pi}_{\text{suite}} = 
\frac{\hat{\pi}_{\text{suite}}(c)}{\sum_{c} \hat{\pi}_{\text{suite}}(c)}.
\]
Integer scenario counts were then obtained as
\[
k(c) = \operatorname{round}\bigl(1600 \cdot \hat{\pi}_{\text{suite}}(c)\bigr),
\quad \text{with} \quad \sum_{c} k(c) = 1600.
\]

The resulting \(\hat{\pi}_{\text{suite}}\) thus constitutes a \emph{dependency-aware empirical distribution}, 
reflecting both realistic correlations among operational factors and deliberate oversampling of safety-critical conditions. 
By design, this distribution differs from the TOD prior \(\boldsymbol{\theta}_{\mathrm{TOD}}\), 
introducing a controlled mismatch that enables the representativeness analysis to quantify how these oversampling strategies 
affect alignment with the expected real-world operational domain.

\add{As shown in Fig.~\ref{fig_represent_process}, the numerical example illustrates three distributions: the empirical scenario-suite distribution obtained directly from the 1600 generated scenarios; the unknown true TOD distribution representing real operational exposure; and the imprecise TOD distribution, given by the interval-valued posterior inferred from limited operational samples and prior assumptions. Representativeness is assessed by comparing the scenario-suite distribution with this uncertain TOD distribution.
}

\subsection{Problem Setting: TOD}
The prior mean vector $\boldsymbol{y}^{(0)}$ over the 32 joint categories was defined assuming \emph{dependencies} among the five operational factors, represented through a Bayesian network with three directed links:
\(
R \rightarrow W,\quad D \rightarrow T,\quad R \rightarrow S.
\)
\begin{comment}
This structure captures realistic relationships in the Target Operational Domain (TOD): 
road type influences both weather exposure and typical speed, while time of day affects traffic conditions.

The root marginal distributions are specified as
\begin{align}
    &P(R) = [0.60,\,0.40] \quad  (\text{Highway, Urban}), \nonumber \\
&P(D) = [0.65,\,0.35] \quad (\text{Day, Night}) \nonumber
\end{align}

The corresponding conditional probability tables (CPTs) are summarized in Table~\ref{tab:cpt_tod}.

\begin{table}[h!]
\centering
\caption{Conditional Probability Tables (dependent TOD prior).}
\vspace{-6pt}
\label{tab:cpt_tod}
\begin{tabular}{lcc}
\hline
\multicolumn{3}{l}{\textbf{$P(W \mid R)$ \; (Weather $\mid$ Road type)}}\\
\hline
 & Clear & Adverse \\
\hline
Highway & 0.75 & 0.25 \\
Urban   & 0.625 & 0.375 \\
\hline
\\[-0.9em]
\hline
\multicolumn{3}{l}{\textbf{$P(T \mid D)$ \; (Traffic $\mid$ Time of day)}}\\
\hline
 & Light & Heavy \\
\hline
Day   & 0.65 & 0.35 \\
Night & 0.50 & 0.50 \\
\hline
\\[-0.9em]
\hline
\multicolumn{3}{l}{\textbf{$P(S \mid R)$ \; (Speed $\mid$ Road type)}}\\
\hline
 & $\le 50$ & $> 50$ \\
\hline
Highway & 0.30 & 0.70 \\
Urban   & 0.65 & 0.35 \\
\hline
\end{tabular}
\end{table}
\end{comment}
Given this dependency structure, the joint prior mean is computed as
\begin{align}
    &y^{(0)}_{(w,r,d,t,s)}
= 
P(R{=}r)\, \times
P(D{=}d)\, \times \nonumber\\
& \hspace{0.3 cm}P(W{=}w \mid R{=}r)\, \times
P(T{=}t \mid D{=}d)\, \times
P(S{=}s \mid R{=}r) \nonumber
\end{align}
Normalization ensures that $\sum_{w,r,t,d,s} y^{(0)}_{(w,r,t,d,s)} = 1$.

The prior strength is modeled as \emph{imprecise},
\[
n^{(0)} \in [\underline{n}, \overline{n}] = [5,\,20],
\]
representing uncertain confidence in the baseline distribution. 
%This imprecision directly determines the lower and upper posterior bounds.

\paragraph{Posterior inference}
Given observed category counts $\mathbf{k} = (k_1,\ldots,k_{32})$, the posterior distribution is
\[
\boldsymbol{\theta}_{\mathrm{TOD}} \mid \mathbf{k}, n^{(0)} 
\sim 
\mathrm{Dirichlet}\!\big(n^{(0)}\mathbf{y}^{(0)} + \mathbf{k}\big),
\]
with posterior mean per category (for illustration purpose):
\[
\mathbb{E}[\theta_k \mid \mathbf{k}, n^{(0)}]
= \frac{n^{(0)}y_k^{(0)} + k_k}{n^{(0)} + n}.
\]
Evaluating this expression at $n^{(0)} = \underline{n}$ and $n^{(0)} = \overline{n}$ yields the \emph{lower and upper posterior expectations}, defining credible bounds on the TOD distribution. %In the numerical examples we have applied $n^{(0)}=10$.

\subsection{Numerical results}
\paragraph{Local representativeness analysis across joint operational categories}
Fig.~\ref{fig_Category_Based_Comparison} shows two probability distributions defined over the same 32 categories:
\begin{figure}[h!]
    \centering
    \includegraphics[width=0.95\linewidth]{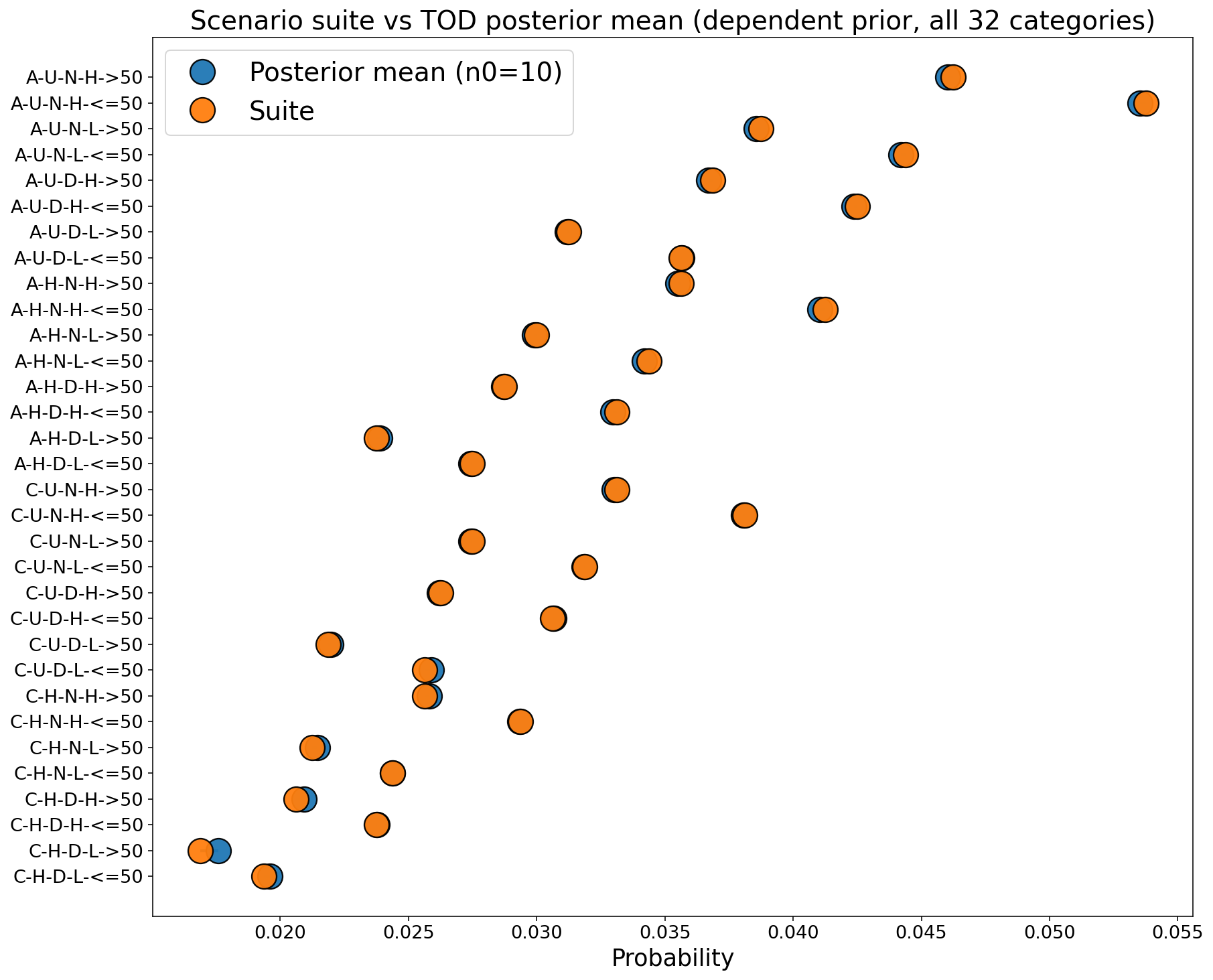}
    \caption{Comparison between the empirical scenario suite distribution and the posterior mean of the TOD distribution (assuming $n^{(0)}=10$).}
    \vspace{-6pt} \label{fig_Category_Based_Comparison}
\end{figure}

Each pair of points represents one unique combination of operational conditions 
(e.g., weather, road type, time of day, traffic, and speed). 
Although this is not a \emph{joint comparison} aggregated across variables, 
it provides a fine-grained, category-level view of how closely the scenario suite aligns with the inferred TOD distribution.

The near overlap between the orange (suite) and blue (posterior TOD) points indicates a strong alignment for most categories. 
Small deviations, typically for adverse or nighttime conditions, reflect deliberate over-sampling in the scenario suite design. 
This visualization highlights which operational categories contribute most to representativeness discrepancies before summarization with global metrics (e.g., TVD, JSD).

%\add{A high discrepancy indicates that the scenario suite does not reflect real operational conditions. Developers can use the per-category differences to identify under-represented and over-represented situations, add or generate scenarios where coverage is insufficient, and reduce redundant scenarios where coverage is excessive. This process can be iterated until the discrepancy falls below an acceptable level.}
\add{A high discrepancy indicates that the scenario suite does not reflect real operational conditions. Per-category differences can identify under- and over-represented situations, guide scenario generation or reduction, and iteratively refine the suite until the discrepancy falls below an acceptable level.
}

\paragraph{Joint-distribution representativeness}
Beyond individual categories, the global similarity between the scenario-suite
distribution $\hat{\boldsymbol{\pi}}_{\mathrm{suite}}$ and the expected posterior TOD
distribution\footnote{Depending on the comparison objective, it is also possible to compare the lower-bound (conservative) and upper-bound distributions.} $\mathbb{E}\!\left[\boldsymbol{\theta}_{\mathrm{TOD}} \mid \mathbf{k},\, n^{(0)}\right]$
is quantified using the TVD~\eqref{eq_TVD} and the JSD~\eqref{eq_JS}.
These measures capture the overall mismatch between the two 32-dimensional
joint distributions, independently of the category ordering.
The values $D_{\mathrm{TV}} = 0.00227$ and $D_{\mathrm{JS}} = 0.00001$ (at $n^{(0)}=10$) indicate that the scenario suite is highly representative of the inferred TOD, with negligible probability redistribution across conditions.

Over the prior-strength interval $n^{(0)} \in [5,\,20]$, 
the distances vary within
$D_{\mathrm{TV}} \in [0.00114,\,0.00452]$ and 
$D_{\mathrm{JS}} \in [0.00000,\,0.00004]$.
The TVD shows an overall mismatch below 0.5\% (Maximum distance $\times100=0.00452 \times 100\simeq 0.5$).
The JSD further shows that the two distributions
carry almost identical information content:
%, with no meaningful differences in their probabilistic profiles. These metrics 
they demonstrate that the scenario suite remains globally representative
of the operational conditions characterized by TOD.

\paragraph{Effect of prior values}
The results in Tables ~\ref{tab_effect_n^0_interval} and~\ref{tab_TV_intervals} quantify how the TVD between the scenario suite and the posterior TOD distribution varies 
with the prior strength parameter $n^{(0)}$ in the imprecise model.

\begin{table}[h!]
\centering
\caption{TVD for different fixed prior-strength values.}
\vspace{-6pt} % reduce space between caption and table
\label{tab_effect_n^0_interval}
\begin{tabular}{c|ccc}
\toprule
$n^{(0)}$ & 5 & 10 & 20 \\ 
\midrule
$D_{\mathrm{TV}}$ & 0.0011 & 0.0023 & 0.0045 \\
\bottomrule
\end{tabular}
\end{table}

%The observed values of $D_{\mathrm{TV}}$ are very small (all below $0.005$), indicating a high level of representativeness according to the interpretative thresholds introduced earlier ($D_{\mathrm{TV}} < 0.05$ corresponds to excellent representativeness). 
As the prior strength $n^{(0)}$ increases, the posterior distribution becomes more influenced by the prior and less by the observed scenario frequencies, leading to a gradual increase in $D_{\mathrm{TV}}$~\ref{tab_effect_n^0_interval}. 
This monotonicity caused by iLUCK model~\cite{imprecision_luck_walter_2009} reflects how epistemic confidence in the prior shifts the posterior away from the empirical scenario suite (small width of the TVD intervals, e.g., $[0.0031, 0.0045]$, for small width of $n^{(0)}$, e.g., $n^{(0)} \in [15,20]$).

%\add{While the numerical example focuses on a scenario suite that is broadly aligned with the inferred TOD, the proposed metrics are not restricted to such positive cases. In the presence of strong mismatch—for example, when scenarios are generated independently of the TOD structure or uniformly at random—the TVD and JSD values would increase toward their upper bounds, clearly indicating poor representativeness. The framework is therefore capable of identifying both well-aligned and highly non-representative scenario suites; the present example is intended to illustrate the interpretation of the metrics rather than to exhaustively enumerate all mismatch cases.}
%\add{While the numerical example considers a scenario suite broadly aligned with the inferred TOD, the proposed metrics are not limited to such cases. Under strong mismatch—for example, when scenarios are generated independently of the TOD structure or uniformly at random—TVD and JSD increase toward their upper bounds, indicating poor representativeness. The framework thus identifies both well-aligned and highly non-representative suites.}
\add{While the numerical example considers a scenario suite broadly aligned with the inferred TOD, the proposed metrics also capture strong mismatch: when scenarios are generated independently of the TOD structure or uniformly at random, TVD and JSD increase toward their upper bounds, indicating poor representativeness and enabling identification of highly non-representative suites.
}

\begin{table}[h!]
\centering
\caption{Effect of the length of the interval of $n^{(0)} \in [a,b]$.}
\vspace{-6pt} % tighten space between caption and table
\label{tab_TV_intervals}
\footnotesize
\begin{tabular}{@{}cccccc@{}}
\toprule
\textbf{$\underline{n}^{(0)}$} & 
\textbf{$\overline{n}^{(0)}$} & 
\textbf{$D_{\mathrm{TV}}(\underline{n}^{(0)})$} & 
\textbf{$D_{\mathrm{TV}}(\overline{n}^{(0)})$} & 
\textbf{$\underline{D}_{\mathrm{TV}}$} & 
\textbf{$\overline{D}_{\mathrm{TV}}$} \\ 
\midrule
5  & 20 & 0.0011 & 0.0045 & 0.0011 & 0.0045 \\
10 & 20 & 0.0023 & 0.0045 & 0.0023 & 0.0045 \\
15 & 20 & 0.0031 & 0.0045 & 0.0031 & 0.0045 \\
\bottomrule
\end{tabular}
\end{table}

\section{Discussion}
\label{sec_discussion_conclusion}
\add{\subsection{Limitations of single-valued representativeness metrics}
%As discussed in Section~\ref{sec_related_work}, existing approaches implicitly assume that the reference distribution is known with sufficient confidence. The analysis above shows why this assumption is problematic in data-sparse settings: a single-valued representativeness metric would conceal sensitivity to prior assumptions, whereas interval-valued results expose the range of conclusions that remain plausible.
As discussed in Section~\ref{sec_related_work}, existing approaches assume the reference distribution is known with sufficient confidence. In data-sparse settings, this is problematic: single-valued representativeness metrics conceal sensitivity to prior assumptions, whereas interval-valued results reveal the range of plausible conclusions.
These are not meant to replace single-valued metrics, but to complement them when data are limited or uncertain by exposing epistemic uncertainty that single-point results would otherwise hide.
}

\subsection{Prior distribution}
\label{sec_difference_dists}
In applying the proposed model, the main task is to specify the prior mean vector $\boldsymbol{y}^{(0)}$ and the prior strength $n^{(0)}$, which together represent the analyst’s belief about the TOD before observing data. Each element of $\boldsymbol{y}^{(0)}$ represents the expected frequency of a TOD condition (for example, how often ``rainy–urban–night'' situations occur), while $n^{(0)}$ quantifies the analyst’s confidence in this belief. %Conceptually, the posterior distribution results from combining the prior belief with new evidence, where $n^{(0)}$ determines the relative influence of the prior compared to the observed data.

%The prior distribution represents the analyst’s baseline knowledge or assumptions about the Target Operational Domain (TOD) before any new data are observed. It is characterized by two components: the prior mean vector $\boldsymbol{y}^{(0)}$, which expresses the expected frequency of each operational condition, and the prior strength $n^{(0)}$, which quantifies how confident we are in those expectations. A higher value of $n^{(0)}$ assigns more influence to the prior in the posterior update, whereas smaller values allow the data to dominate.

In practice, the prior mean $\boldsymbol{y}^{(0)}$ is not fixed but inferred from various forms of background knowledge. When sufficient operational or simulation data are available, the empirical proportions of observed conditions can be used directly as an estimate of $\boldsymbol{y}^{(0)}$. In more data-sparse contexts, expert elicitation provides an alternative: domain specialists may indicate how often specific conditions are likely to occur (for example, “most operations occur on highways during daytime”). These qualitative statements are translated into numerical probabilities that sum to one. In cases where experts can only specify approximate ranges rather than precise probabilities, these ranges define upper and lower bounds for the corresponding prior mean components.

In addition, published reference statistics—such as national traffic surveys, meteorological archives, or the multi-source environmental datasets used in Hiller et al.~\cite{hiller_creating_2022}—can provide baseline frequency estimates for relevant factors like weather type, road category, or traffic density. When no credible information exists, a uniform prior $\boldsymbol{y}^{(0)} = (1/K,\ldots,1/K)$ expresses a state of complete ignorance by assigning equal probability to all possible categories.

\subsection{Dependence and Independence among TOD Categories}
In this study, the TOD is defined as a combination of discrete operational factors such as weather, road type, illumination, and traffic density etc. 
To derive the joint distribution, it is necessary to account for the dependencies that may exist among these factors, as such factors are rarely independent in real-world operational environments. 
Certain weather conditions may be more frequent on rural roads than in urban areas; night-time driving is typically associated with lower traffic density; and fog or heavy rain may be confined to specific geographical regions or seasons. 
These dependencies imply that the true TOD distribution $\boldsymbol{\theta}_{\mathrm{TOD}}$ cannot, in general, be expressed as a simple product of marginal probabilities:
\[
P(\text{Road}, \text{Weather}, \text{Time}) \neq P(\text{Road}) \, P(\text{Weather}) \, P(\text{Time}).
\]
Neglecting such correlations can lead to systematic under- or over-representation of certain operational categories in the scenario suite, biasing the representativeness assessment.

In Sec.~\ref{sec_numerical_example}, dependencies were explicitly modeled to reflect realistic operational conditions. 
These dependencies need to be determined based on prior expert knowledge and supported by available historical data, which indicate systematic relationships between certain environmental and contextual factors. If the dependencies are not well understood and independence between variables must be assumed, a sensitivity analysis can help to avoid misleading conclusions by quantifying the potential impact of this assumption on the representativeness metrics.  

\section{Summary and Conclusions}
\label{sec_conclusion}

This paper presented a probabilistic framework for assessing the \emph{representativeness} of scenario datasets with respect to the TOD. 
%%Building upon the definition of representativeness in ISO~8800~\cite{ISO8800_2024}, 
Unlike existing approaches based on
fixed distributions and single-point discrepancy measures, the proposed approach models the TOD as an uncertain categorical distribution over joint operational conditions and applies a robust Bayesian formulation with imprecise Dirichlet priors to capture epistemic uncertainty about both prior beliefs and data sufficiency. 
The resulting posterior \emph{credal set} enables transparent quantification of uncertainty in representativeness metrics (e.g., Total Variation Distance) between the observed scenario suite distribution and the inferred TOD distribution. 

\add{%The numerical example uses a small set of abstract scenario features to clearly illustrate the proposed concept. 
Unlike the numerical example, real-world ODDs involve many variables; the framework therefore relies on structured or factorized representations capturing only relevant dependencies.}%\footnote{In practice, the method focuses on relevant feature groups and relationships instead of modeling all possible feature combinations.}
%. Bayesian inference can then be performed using limited or aggregated operational data, with uncertainty naturally decreasing as more data become available.}

Future work will extend the current framework along several directions. 
\add{First, we will relax the i.i.d. assumption to account for potential correlations between scenarios, which may arise from temporal, spatial, or operational dependencies in real-world driving data.}
Next, we will address the dependence between the TOD and scenario suite distributions, acknowledging that both often stem from the same limited real-world dataset. Treating them as fully independent can conceal shared biases and limit the ability to reveal representational gaps. 
We also aim to relax the simplifying assumptions made in this study by incorporating continuous operational factors, thereby moving beyond purely discretized representations toward a hybrid framework that combines discrete and continuous variables. 
Beyond representativeness, we will develop quantitative metrics for other data-related safety properties—such as completeness, consistency, and relevance—and explore their interrelations within a unified framework. 
Ultimately, these assessments should be embedded within an \emph{Assurance~2.0} case (e.g., \cite{Chen_assurance2_2025}), linking probabilistic evidence to structured, auditable safety arguments for trustworthy AI. %\add{Finally, while the numerical example demonstrates the feasibility and interpretability of the approach, a broader empirical comparison against existing representativeness metrics across larger and more realistic datasets is left for future work.}
\add{While the numerical example demonstrates the feasibility and interpretability of the approach, broader empirical comparisons with existing representativeness metrics on larger and more realistic datasets are left for future work.}

\section*{Acknowledgment}
This work is supported by the European Union's EU Framework Program for Research and Innovation Europe Horizon (grant agreement No 101202457) and Innovate UK through the DriveSafeAI Project (10062594). XZ's contribution is supported by the UK EPSRC New Investigator Award [EP/Z536568/1]. SK's contribution is supported by the UKRI Future Leaders Fellowship Grant [MR/S035176/1]. Views and opinions expressed are those of the authors only and do not necessarily reflect those of the European Union or European Research Executive Agency. Neither the European Union nor the granting authority can be held responsible for them.

%The preferred spelling of the word ``acknowledgment'' in America is without  an ``e'' after the ``g''. Avoid the stilted expression ``one of us (R. B. G.) thanks $\ldots$''. Instead, try ``R. B. G. thanks$\ldots$''. Put sponsor acknowledgments in the unnumbered footnote on the first page.

\appendices

\section{Proof Sketch of Theorem~\ref{thm_TOD_posterior}}
\label{proof_outline_appendix}
%\begin{proof}
\paragraph{Model setup}
Let $Z\in\{1,\dots,K\}$ denote the TOD category with probability vector 
$\boldsymbol{\theta}=(\theta_1,\ldots,\theta_K)\in\Delta_{K-1}$.
Given $n$ i.i.d.\ observations with count vector $\mathbf{k}=(k_1,\ldots,k_K)$, 
the likelihood is multinomial:

%\[
%p(\mathbf{k}\mid\boldsymbol{\theta})
%\;\propto\;
%\prod_{k=1}^K \theta_k^{\,k_k}.
%\]
%The sufficient statistic for $\boldsymbol{\theta}$ is $\mathbf{k}$.
We assume that the TOD variable 
$Z$ takes one of $K$ mutually exclusive categories, with probabilities 
$\boldsymbol{\theta} = (\theta_1, \ldots, \theta_K)$, where 
$\sum_{k=1}^{K} \theta_k = 1$.
If we observe $n$ independent samples 
$z_1, \ldots, z_n$ from this categorical distribution and record 
the number of occurrences in each category as 
$\mathbf{k} = (k_1, \ldots, k_K)$, 
then the joint probability of the observed counts follows the 
multinomial distribution:
\[
p(\mathbf{k} \mid \boldsymbol{\theta}) 
= \frac{n!}{\prod_{k=1}^{K} k_k!} 
\prod_{k=1}^{K} \theta_k^{\,k_k},
\qquad \text{with } \sum_{k=1}^{K} k_k = n.
\]
Since the factorial term does not depend on $\boldsymbol{\theta}$, 
it can be dropped when forming the likelihood function for inference:
\[
L(\boldsymbol{\theta} \mid \mathbf{k}) 
\propto \prod_{k=1}^{K} \theta_k^{\,k_k}.
\]
Thus, the multinomial likelihood correctly represents the sampling process for counts over multiple mutually exclusive categories.
The sufficient statistic for $\boldsymbol{\theta}$ is indeed the count 
vector $\mathbf{k}$, since (by the factorization theorem) all information 
in the sample relevant to $\boldsymbol{\theta}$ is contained in these 
category counts.

\paragraph{Dirichlet prior and iLUCK reparameterization.}
Assume a Dirichlet prior 
$\boldsymbol{\theta} \sim \mathrm{Dirichlet}(\boldsymbol{\alpha})$
with parameters $\boldsymbol{\alpha} = n^{(0)}\boldsymbol{y}^{(0)}$,  
where $\boldsymbol{y}^{(0)}$ is the prior mean and $n^{(0)} > 0$ is the prior strength 
(equivalent sample size).  
The Dirichlet distribution is the conjugate prior of the multinomial likelihood.
That means if the likelihood of the data given the parameters is multinomial,
then the posterior distribution over the parameters has the same functional form (Dirichlet),
which allows closed-form Bayesian updating.
This is the iLUCK reparameterization, satisfying  
$\mathbb{E}[\theta_k] = y^{(0)}_k$ and $\alpha_k = n^{(0)} y^{(0)}_k$.

\paragraph{Precise Bayesian update}
By conjugacy of the Dirichlet–multinomial pair,
\[
p(\boldsymbol{\theta}\mid\mathbf{k})
\;\propto\;
p(\mathbf{k}\mid\boldsymbol{\theta})\,p(\boldsymbol{\theta})
\;\propto\;
\prod_{k=1}^K \theta_k^{\,(\alpha_k^{(0)} + k_k - 1)}.
\]
Hence, $\boldsymbol{\theta}_{\text{TOD}}\mid\mathbf{k}
\;\sim\;
\mathrm{Dirichlet}\!\big(n^{(0)}\boldsymbol{y}^{(0)}+\mathbf{k}\big),$
%\[
%\boldsymbol{\theta}_{\text{TOD}}\mid\mathbf{k}
%\;\sim\;
%\mathrm{Dirichlet}\!\big(n^{(0)}\boldsymbol{y}^{(0)}+\mathbf{k}\big),
%\]
and each component marginally follows:
$\theta_k \!\mid \! \mathbf{k} \sim \mathrm{Beta}(a_k, a_0 - a_k)$
with $a_k=n^{(0)}y^{(0)}_k+k_k$ and $a_0=n^{(0)}+n$.

\paragraph{Imprecise Bayesian posterior.}
Represent prior uncertainty by the set  
$\boldsymbol{y}^{(0)}\in\mathcal{Y}_{\text{TOD}}\subseteq\Delta_{K-1}$ and 
$n^{(0)}\in[\underline{n},\overline{n}]$.
By conjugacy, each prior in this set yields a corresponding Dirichlet posterior, forming the posterior credal set:
\begin{align}
&\mathcal{P}(\boldsymbol{\theta}_{\text{TOD}}\mid\mathbf{k})
= \nonumber\\
& \hspace{0.5cm} \big\{
\mathrm{Dirichlet}\!\big(n^{(0)}\boldsymbol{y}^{(0)}+\mathbf{k}\big)
:\,
\boldsymbol{y}^{(0)}\in\mathcal{Y}_{\text{TOD}},\,
n^{(0)}\in[\underline{n},\overline{n}]
\big\} \nonumber
\end{align}
Because the Dirichlet update is affine in $(\boldsymbol{y}^{(0)},n^{(0)})$, the posterior set remains convex.

\paragraph{Posterior envelopes and extreme points}
%The prior uncertainty induces a \emph{continuum of Dirichlet posterior distributions}
%\[
%\mathcal{P}(\boldsymbol{\theta}_{\text{TOD}} \mid k)
%= 
%\big\{
%\mathrm{Dirichlet}\big(n^{(0)} y^{(0)} + k\big)
%:\;
%y^{(0)} \in \mathcal{Y}_{\text{TOD}},\;
%n^{(0)} \in [\underline{n}, \overline{n}]
%\big\},
%\]
%which together form a \emph{posterior credal set} over the simplex 
$\Delta_{K-1}$.
Under the iLUCK~\cite{imprecision_luck_walter_2009} monotonicity result,
the lower and upper bounds of this joint posterior family (the \emph{posterior envelope})
are attained at the \emph{extreme points} of 
$\mathcal{Y}_{\text{TOD}} \times [\underline{n}, \overline{n}]$.
Monotonicity holds because, under the iLUCK reparameterization,
the Dirichlet shape parameters
\[
a_k = n^{(0)} y_k^{(0)} + k_k, 
\qquad 
a_0 = n^{(0)} + n,
\]
depend \emph{affinely} on the prior mean and strength.
This affine relationship preserves convexity of the credal set under updating,
ensuring that posterior expectations, densities, and CDFs vary monotonically
with $y^{(0)}$ and $n^{(0)}$.
Consequently, the lower and upper \emph{Dirichlet posterior densities}—
that is, the joint posterior envelope over $\Delta_{K-1}$—
are obtained by evaluating the Dirichlet density at the extreme prior configurations.
Marginally, each component $\theta_k$ follows a Beta distribution whose CDF
is the regularized incomplete Beta function $I_t(a_k, a_0 - a_k)$,
so the same boundary evaluation yields tight marginal bounds on posterior beliefs.
The Beta marginal is correct because each component of a Dirichlet distribution
has a univariate Beta distribution when integrating out the remaining dimensions,
preserving exact consistency between the joint and marginal posterior forms.
Hence, the marginal posterior envelopes can be estimated \emph{in the same way},
by evaluating the Beta distributions at the same extreme prior configurations.

	\bibliographystyle{IEEEtran}
	% Your .bib file here
	\bibliography{root} 
	
\end{document}

%%%%%%%%%%%%%%%%%%%%%%%%%%%%%%%%%%%%%%%%%%%%%%%%%%%%%%%%%%%%%%%%%%